\documentclass[letterpaper]{article} 
\usepackage[submission]{aaai24}  
\usepackage{times}  
\usepackage{helvet}  
\usepackage{courier}  
\usepackage[hyphens]{url}  
\usepackage{graphicx} 
\usepackage{natbib}  
\usepackage{caption} 
\usepackage{algorithm}
\usepackage{algorithmic}
\usepackage{utfsym}
\usepackage{amsmath}
\usepackage{newfloat}
\usepackage{listings}
\DeclareCaptionStyle{ruled}{labelfont=normalfont,labelsep=colon,strut=off} 
\floatstyle{ruled}
\newfloat{listing}{tb}{lst}{}
\floatname{listing}{Listing}
\title{EA-LSS: Edge-aware Lift-splat-shot Framework for 3D BEV Object Detection}
\author{
    Haotian Hu\textsuperscript{\rm 1}, Fanyi Wang\textsuperscript{\rm 2}\equalcontrib, Jingwen Su\textsuperscript{\rm 2}, Yaonong Wang\textsuperscript{\rm 1}, Laifeng Hu\textsuperscript{\rm 1} \\ Weiye Fang\textsuperscript{\rm 1}, Jingwei Xu\textsuperscript{\rm 1}, Zhiwang Zhang\textsuperscript{\rm 3}\equalcontrib \\
    \small
    \textsuperscript{\rm 1} Zhejiang Leapmotor Technology CO., LTD.\\
    \textsuperscript{\rm 2} OPPO Research Institute\\
    \textsuperscript{\rm 3} NingboTech University\\
    \{hu\_haotian,wang\_yaonong, hu\_laifeng, fang\_weiye, xu\_jingwei\}@leapmotor.com, \\ \{wangfanyi,sujingwen\}@oppo.com,\\  zhiwang.zhang@nit.zju.edu.cn
}

\begin{document}

\maketitle


\begin{abstract}
   In recent years, great progress has been made in the Lift-Splat-Shot-based (LSS-based) 3D object detection method.
   However, inaccurate depth estimation remains an important constraint to the accuracy of camera-only and multi-modal 3D object detection models, especially in regions where the depth changes significantly (i.e., the ``depth jump'' problem).
   In this paper, we proposed a novel Edge-aware Lift-splat-shot (EA-LSS) framework. Specifically, edge-aware depth fusion (EADF) module is proposed to alleviate the ``depth jump'' problem and fine-grained depth (FGD) module to further enforce refined supervision on depth. 
   Our EA-LSS framework is compatible for any LSS-based 3D object detection models, and effectively boosts their performances with negligible increment of inference time. 
   Experiments on nuScenes benchmarks demonstrate that EA-LSS is effective in either camera-only or multi-modal models. It is worth mentioning that EA-LSS achieved the \textbf{state-of-the-art} performance on nuScenes test benchmarks with mAP and NDS of 76.5\% and 77.6\%, respectively.
\end{abstract}

\section{Introduction}

With the development of autonomous driving, vehicle sensors become more complex. Integrating multi-source information from different sensors (e.g., 2D images, 3D lidar, and radar) and characterizing their features in a unified way becomes crucial. Projection of various features from various sources into Bird's-Eye-View (BEV)~\cite{bai2022transfusion,huang2021bevdet,li2022bevdepth,liu2022bevfusion} becomes prevalent, and has attracted a great deal of attentions from academia and industry. One of the core problems of BEV perception task is how to reconstruct the hidden depth information in 2D images, and provide accurate BEV features for subsequent networks(i.e., BEV decoder and detection head).

\begin{figure}[t]
\begin{center}
\includegraphics[width=\linewidth]{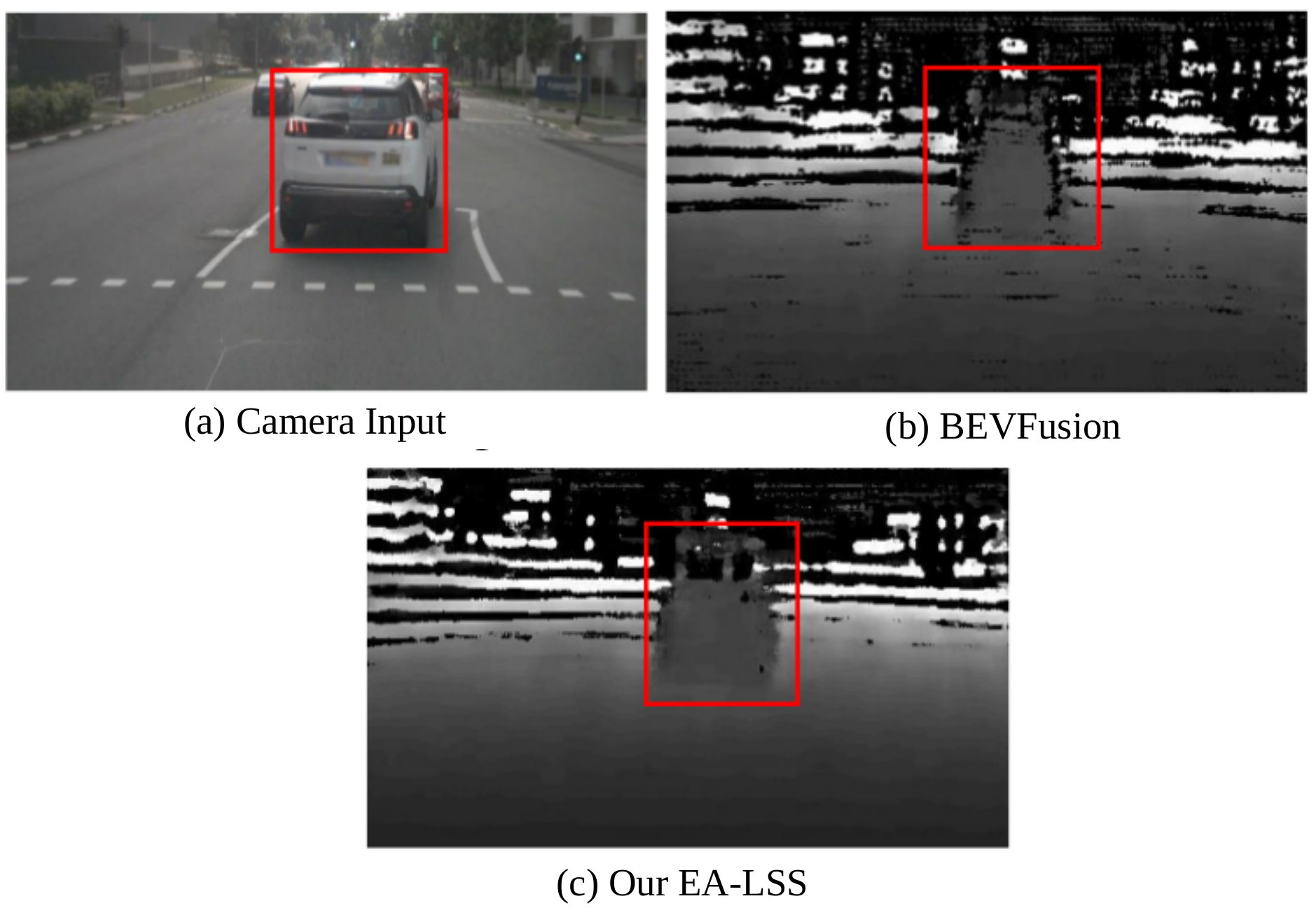}
\end{center}
   \caption{Visualisation of ``depth jump'' problem. It shows depth estimation before and after adding EA-LSS framework on baseline method BEVFusion~\cite{liu2022bevfusion}. The larger the pixel value, the deeper the depth. }
   \label{fig:motivation}
\end{figure}

Lift-Splat-Shot (LSS)~\cite{philion2020lss} predicts the depth distribution of each pixel on 2D feature map, and ``lifts'' 2D features of each mesh into voxel space by the corresponding depth estimation. However, due to the large depth difference (i.e., ``depth jump'') among regions in the real scene, 
depth estimate network of existing LSS-based methods~\cite{huang2021bevdet, li2022bevdepth, liang2022bevfusion, liu2022bevfusion} are unable to derive accurate depth estimation.
And it results in 2D image features at the edges of the scene ``lifted'' to wrong voxel space. As shown in Figure~\ref{fig:motivation}, the black area around edges of the foreground vehicle (shown in red box) indicates poor performance of depth estimation network in baseline method BEVFusion~\cite{liang2022bevfusion}, when dealing with the ``depth jump" region.

In order to improve the accuracy of depth estimation networks in the ``depth jump" region, we propose an edge-
aware depth fusion (EADF) module. It provides additional information about the scene edges to the depth network, and helps the model better adapt to the rapid changes in depth among objects.

We find that existing LSS-based methods did not fully utilize depth information when matching the predicted depth map and the ground-truth depth map, leading to poor performance of the depth network. BEVDet~\cite{huang2021bevdet} does not consider using depth information of point cloud to constrain the depth network. BEVFusion~\cite{liang2022bevfusion} uses the depth of point cloud as a feature to implicitly supervise the depth estimation network, which indicates that accurate depth information is not fully utilized. CaDDN~\cite{CaDDN} interpolates sparse depth map projected of LIDAR points, and uses it to supervise prediction of depth distribution. But the interpolated depth ground-truth tends to be smooth and loses the information of the depth distribution between the foreground and the background. BEVDepth~\cite{li2022bevdepth} utilizes the minimum pooling and one hot operations to align the depth estimates and the projected depth of point cloud. But depth of point cloud is extremely sparse when projected to 2D, which hinders the network from learning the correct depth estimates.

To solve the above problem, we propose a novel fine-grained depth (FGD) module. This module contains an upsampling branch to match the size between the predicted feature map and the ground-truth depth map during training. It allows the depth estimation network to perceive the depth distribution of the whole scene in a finer way, and preserves the original depth information as much as possible. The proposed FGD module is only used in training stage, so it does not affect the speed and resource consumption of the model during inference stage.

Our proposed Edge-aware Lift-splat-shot (EA-LSS) framework couples the edge-aware depth fusion (EADF) module and the fine-grained depth (FGD) module. As a plug-and-play view converter, EA-LSS framework can be adapted to various LSS-based BEV 3D object detection methods~\cite{huang2021bevdet,liang2022bevfusion, CaDDN, li2022bevdepth}. Our proposed EA-LSS framework can assist the depth estimation network in estimating the depth distribution for the monocular image in a finer way, and adapt to the ``depth jump" region in the image in a better way. 

We validate our proposed EA-LSS framework on nuScenes~\cite{caesar2020nuScenes} 3D object detection benchmark. Our method improves the mAP and NDS of the camera-only baseline Tig-bev~\cite{huang2022tigbev} by 2.1\% and 3.2\%, the multi-modal baseline BEVFusion~\cite{liu2022bevfusion} by 1.6\% and 1.0\%, with negligible increase on time and computation resources in the inference stage.  In the nuScenes test dataset, our proposed EA-LSS framework achieves the state-of-the-art performance, with mAP and NDS of 76.5\% and 77.6\%, respectively.

Contributions of this paper are summarized as follows:
\begin{itemize}
    \item An edge-aware depth fusion (EADF) module and a fine-grained depth (FGD) module are proposed, which effectively alleviate the problem of weak fitting ability of depth estimation networks to regions with rapid changes in image depth, and misalignment of depth prediction and truth value during size matching.
    \item A novel Edge-aware Lift-splat-shot (EA-LSS) framework is proposed, as a multi-modal depth prediction paradigm, effectively utilizes depth information for depth estimation tasks.
    \item Comprehensive experiments demonstrate that with negligible increment of inference time and resources, EA-LSS significantly improves performance of several state-of-the-art BEV baselines on the nuScenes 3D object detection benchmark. Our proposed EA-LSS framework achieves the top 1 in nuScenes detection task learderboard 
    with mAP and NDS of 76.5\% and 77.6\%, respectively.
\end{itemize}

\section{Related Work}
3D object detection is one of the core tasks of 3D perception that predicts objects of specific types and 3D bounding boxes. Previous works~\cite{zhou2019centernet, wang2021fcos3d,lang2019pointpillars,shi2019pointrcnn,yan2018second,chen2022futr3d,vora2020pointpainting} directly predict from perspective view (PV) features, which are unable to exploit stereo visual cues from multiple views and time-continuous frames. BEV-based 3D object detection models~\cite{liu2022bevfusion, liang2022bevfusion, reading2021categorical, huang2021bevdet,huang2022bevdet4d,yin2021Multimodal} excel at fusing multi-source and multi-timestamp features, and have made great progress in terms of efficiency and performance.

One of the core problems of BEV perception lies in reconstructing the lost 3D information using PV-to-BEV view translation. To compensate the difference between these two views, previous methods can be divided into network-based view transformation methods~\cite{li2021hdmapnet, lu2019monocular, li2022bevformer, wang2022detr3d, liu2022petrv2} and depth-based view transformation methods~\cite{huang2021bevdet, reading2021categorical, you2020pseudolidar++, wang2022mvfcos3d++, wu2022sfd}. 

\subsection{Network-based View Transformation Methods}\label{subsection:network-based}
\subsubsection{MLP-based Approach.}
MLP-based approaches use MLP to learn an implicit representation of camera calibration for conversion between two different views. HDMapNet~\cite{li2021hdmapnet} considers that one-way projection is difficult to ensure effective transfer during forward pass of view information from PV to BEV. Therefore, it also uses MLP to backward project features from BEV to PV. VED~\cite{lu2019monocular} is the first method using end-to-end learning on monocular images to generate semantic-metric occupancy grid map in real time. It converts PV feature maps into BEV feature maps through flattening, mapping, and reshaping operations for multi-modal perception and prediction.

\subsubsection{Transformer-based Approach.}
Transformer-based approaches directly employ the transformer to map PV into BEV, which does not explicitly require the camera model. Tesla first uses the transformer to project PV features into BEV. It adopts positional encoding to design a set of BEV queries, then performs view transformations through cross-attention mechanism between BEV queries and image features.  BEVFormer~\cite{li2022bevformer} employs deformable attention to extract dense queries in BEV and describes relationships between multi-view features. DETR3D~\cite{wang2022detr3d} utilizes a geometry-based feature sampling process instead of cross-attention to predict 3D reference points. And reference points are projected into the 2D image plane using a calibration matrix to achieve end-to-end 3D edge prediction. PETRv2~\cite{liu2022petrv2} extends 3D position embedding to the time domain, effectively utilizing continuous frame information. PolarDETR~\cite{chen2022polar} brings parameters of 3D object detection to the polar coordinate system, reformulating edge parameterization, network prediction and loss calculation.
\begin{figure*}[t]
\begin{center}
\includegraphics[width=\linewidth]{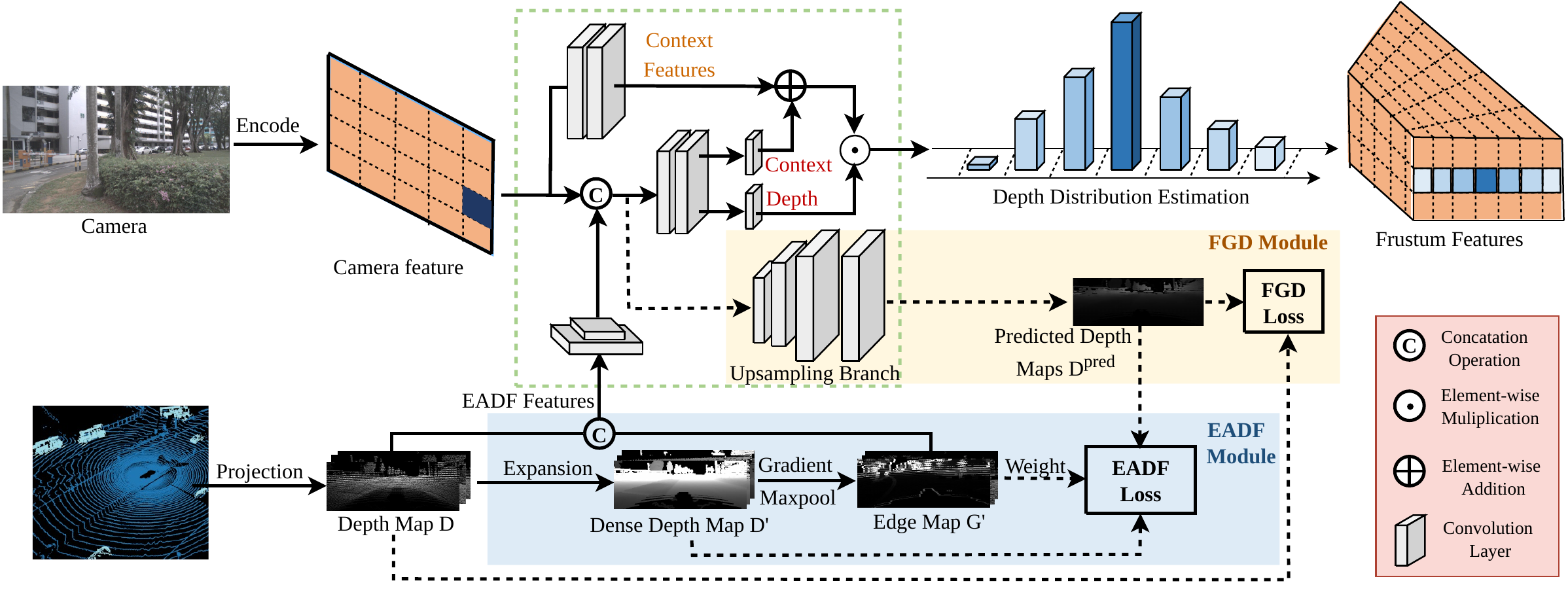}
\end{center}
   \caption{The details of our proposed EA-LSS Framework. It contains an edge-aware depth fusion (EADF) module (shown in blue) and a fine-grained depth (FGD) module (shown in yellow). In this framework, the multi-view 2D camera images and 3D Lidar point clouds are used as input to generate the frustum features. The dash lines are only involved in inference stage. } 
   \label{fig:EABEV_architecture}
\end{figure*}

\subsection{Depth-based View Transformation} \label{subsection:depth-based}
\subsubsection{Point-based Approach.}
Point-based approaches use depth estimation networks to convert image pixels into pseudo-laser point clouds, which are fed to a lidar-based 3D detector. These approaches are highly dependent on the accuracy of depth estimation. Pseudo-lidar++~\cite{you2020pseudolidar++} improves the accuracy of depth estimation network using a stereo vision estimation network. SFD~\cite{wu2022sfd} uses both the original point cloud and the pseudo point cloud generated by depth estimation network, and fuses them to improve detection accuracy of distant and occluded objects.

\subsubsection{Depth Estimation-based Approach.}
Depth estimation-based methods explicitly predict depth distribution of image features to construct 3D features. LSS~\cite{philion2020lss} is a representation of these methods. LSS-based methods predict both feature distribution in the depth direction and image context information, to determine the features of each point of the perspective ray, reducing the loss of image semantics due to depth prediction bias.
BEVDet~\cite{huang2021bevdet} proposes a multi-view 3D detection framework that includes an image view encoder, view transformer, BEV encoder, and detection head. CaDDN~\cite{reading2021categorical} interpolates the sparse depth map of lidar point projection and supervises the depth estimation using depth distribution. 
MV-FCOS3D++~\cite{wang2022mvfcos3d++} proposes the pre-training of depth estimation and monocular 3D detection, which significantly enhances the learning ability of 2D backbones.

Our proposed EA-LSS framework explores refinement of depth estimation-based approach by the proposed depth estimation module and edge-aware depth fusion module.
EA-LSS framework is able to accurately estimate the global depth and the edge regions of objects with sharp depth changes.  
And ``depth jump'' problem is effectively alleviated, which is meaningful to properly guide subsequent network.

\section{Method}
Depth estimation network is crucial in both camera-only and multi-modal 3D object detection methods. Estimation accuracy can seriously affect performance of the subsequent network (i,e., BEV decoder and detection head). Therefore, in order to predict the fine-grained depth distribution, we propose the edge-aware lift-splat-shot (EA-LSS) framework (Figure~\ref{fig:EABEV_architecture}) coupled by an edge-aware depth fusion (EADF) module and a fine-grained depth (FGD) module.  


\subsection{Fine-grained Depth Module}\label{subsection:FGD}
We propose a fine-grained depth module to constrain the depth network pixel by pixel. 
In order to retain accurate depth information, an upsampling branch is proposed as additional depth estimation network for supervision. 
The depth map projected from point cloud is extremely sparse. If the loss is calculated between the projected depth map from the point clouds and the predicted depth map, excessive zero values in the projected depth map can increase difficulty of fitting these two depth maps. We propose a fine-grained depth (FGD) loss as constraint to supervise the loss between the non-zero pixels in the projected depth map and their corresponding pixels in the predicted depth map. The proposed FGD loss effectively excludes the interference of the zero-valued pixels of the projected depth map on the depth network. 
\begin{figure}[t]
\begin{center}
\includegraphics[width=\linewidth]{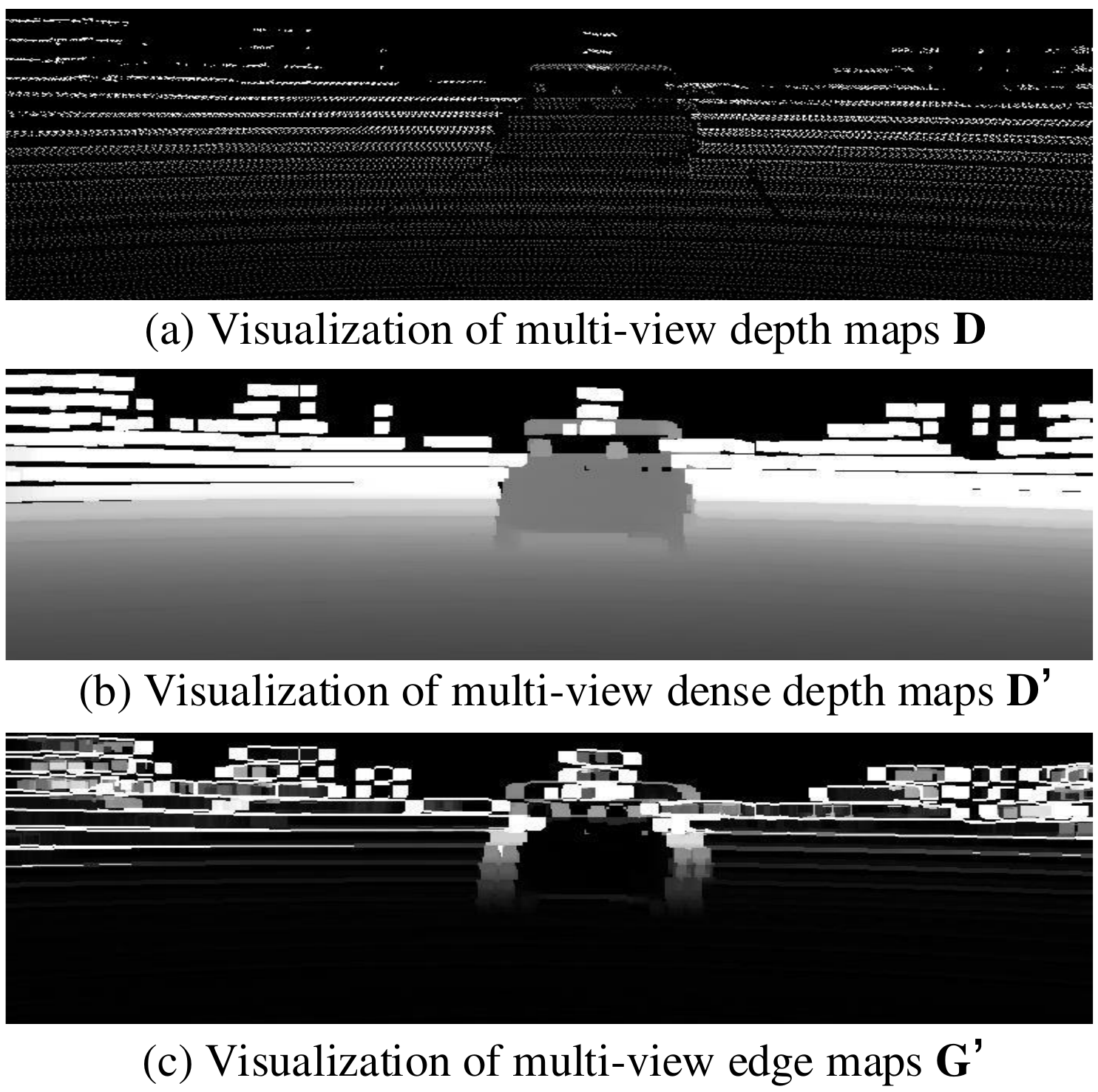}
\end{center}
   \caption{Visualization of (a) multi-view depth maps $\mathbf{D}$, (b) multi-view dense depth maps $\mathbf{D'}$, and (c) multi-view edge maps $\mathbf{G'}$. }
\label{fig:depth_map}
\end{figure}

In addition, to focus on the foreground objects (considering more objects appear in background), focal loss~\cite{focalloss} is embedded in fine-grained depth loss. In FGD loss, the outputs of the sampling branch $\mathbf{D}^{pred}$ are used as predicted samples while the multi-view depth maps $\mathbf{D}$ projected from input point clouds are used as ground-truth labels.  FGD loss is defined below:
\begin{equation}
\label{eq:SDL}
L_{FGD}= { \sum_{i=1}^{n} \sum_{c=1}^{H_D} -\alpha_c  ( 1- {y}_{i,c}   )^{\gamma }log(\widehat{y}_{i,c})}, \
\end{equation}
where $n$ is the number of non-zero pixels in ground-truth $\mathbf{D}$. $H_D$ denotes the classes of predicted depth maps $\mathbf{D}^{pred}$. Here ground-truth $\mathbf{D}$ is transferred to one-hot vectors. $\widehat{y}_{i,c}$ is the value of $c^{th}$ class of $i^{th}$ one-hot vector of non-zero pixels in $\mathbf{D}$.  $y_{i,c}$ is from the corresponding pixels in predicted depth map $\mathbf{D}^{pred}$.
$\alpha_c$ denotes the weight and  $\gamma$ denotes the predefined hyper-parameter for focal loss.

\subsection{Edge-aware Depth Fusion Module}
In order to alleviate the ``depth jump'' problem, we propose an edge-aware depth fusion (EADF) Module.

The input point clouds are projected to multi-view depth maps $ \mathbf{D}=\{\mathbf{d}_{ij}|i=1,...,N_H; j=1,...,N_W\} \in R^{N_v*N_H*H_W}$, where $N_v$ represents the number of views, $N_H$ and $N_W$ represent the length and height of depth maps. 
Multi-view depth maps $\mathbf{D}$ are divided into blocks in size $k*k$ along x-axis and y-axis, where $k$ is the step size. Then we use the maximum depth value of each block to fill the whole block through expanding operation, which connects most scenes in the point cloud depth map. After block division and expanding operation,  multi-view depth maps $\mathbf{D}$ are transferred to the multi-view dense depth maps $\mathbf{D'}$ with the same dimension.

Next, we calculate gradient of multi-view dense depth map along x-axis and y-axis to extract edge-aware 3D geometry information. Considering each axis has two directions, we define gradients of dense depth map $\mathbf{G}=\{\mathbf{G}^1,\mathbf{G}^2,\mathbf{G}^3,\mathbf{G}^4\}$ in Equation \ref{eq:5}, where $\mathbf{G}^t=\{ \mathbf{g}^t_{i,j}\}, t=1,2,3,4$. Each pixel of multi-view dense depth maps $\mathbf{g}^1_{i,j}$, $\mathbf{g}^2_{i,j}$, $\mathbf{g}^3_{i,j}$, and $\mathbf{g}^4_{i,j}$ are defined as:
\begin{equation} \label{eq:5}
\left\{
\begin{array}{llll}
\mathbf{g}^1_{i,j} &=\mathbf{d'}_{i,j} - \mathbf{d'}_{i+k,j},  \\ 
\mathbf{g}^2_{i,j} &=\mathbf{d'}_{i,j} - \mathbf{d'}_{i-k,j},  \\
\mathbf{g}^3_{i,j} &=\mathbf{d'}_{i,j} - \mathbf{d'}_{i,j+k},  \\
\mathbf{g}^4_{i,j} &=\mathbf{d'}_{i,j} - \mathbf{d'}_{i,j-k},  \\
\end{array}\right.
\end{equation}
where $\mathbf{d'}_{i,j}$ is the $i^{th}$ column $j^{th}$ row of  multi-view dense depth maps $\mathbf{D'}$. $\{\mathbf{g}^t_{i,j}\}\in R^{N_v*N_H*H_W}$ represents one direction of the gradient of multi-view dense depth map.

Dense depth map gradient $ \mathbf{G} \in R^{N_v*N_H*H_W*4}$ is passed to the max pooling operation on the last dimension and normalization operation to obtain the multi-view edge maps $\mathbf{G'}$. 
Multi-view edge maps $\mathbf{G'} \in R^{N_v*N_H*H_W}$ represents the edge of different objects. $\mathbf{G'}$ are scaled to $\left[ 0,1 \right] $ by normalization operation.

As shown in Figure~\ref{fig:depth_map}~(a), multi-view depth maps $\mathbf{D}$ projected from 3D lidar scanning are sparse, which is not suitable for directly supervising the depth estimation networks. Figure~\ref{fig:depth_map}~(b) shows the multi-view dense depth maps $\mathbf{D'}$ after block division and expanding operations. Original depth information from depth map is reserved. Main objects in dense depth map are connected, which is beneficial to depth estimation network of the whole scene. In Figure~\ref{fig:depth_map}~(c), we obtain multi-view edge maps $\mathbf{G'}$ by calculating the maximum gradient value of each pixel along the four directions. Multi-view edge maps $\mathbf{G'}$ represent the maximum depth variation among each block. 
After multi-view depth map $\mathbf{D}$ and multi-view edge map $\mathbf{G'}$ are obtained, we concatenate $\mathbf{D}$ and $\mathbf{G'}$ to generate the output of EADF module as $\mathbf{F}^{EADF}= [\mathbf{D}:\mathbf{G'}]$.

We propose edge-aware depth fusion (EADF) loss to focus on the edge of each object, by using multi-view dense depth maps $\mathbf{D'}$ as ground-truth and multi-view edge maps $\mathbf{G'}$ as weight to alleviate the ``depth jump'' problem. EADF
 loss $L_{EADF}$ for supervision is defined:
\begin{equation}
\label{eq:GEDL}
\begin{split}
L_{EADF} &= \sum_{i=1}^{m} \sum_{c=1}^{H_D} -{\alpha_c}\left ( 1-{p}_{i,c}  \right )   ^{\gamma } log\left ( \widehat{p}_{i,c}  \right ) \ast w_{i}, \
\end{split}
\end{equation}
where $m$ is the number of pixels in predicted depth map $\mathbf{D}^{pred}$. $H_D$ denotes the classes of predicted depth. ${p}_{i,c}$  denotes the value of $c^{th}$ class of the $i^{th}$ pixel from predicted depth map $\mathbf{D}^{pred}$, $\widehat{p}_{i,c}$ is  the corresponding pixel in ground-truth depth map $\mathbf{D'}$,  and $w_{i}$ is the corresponding pixel value of $\mathbf{G'}$. $\alpha_c$ denotes the weight and  $\gamma$ denotes the predefined hyper-parameter for focal loss.

\begin{figure}[t]
\begin{center}
\includegraphics[width=\linewidth]{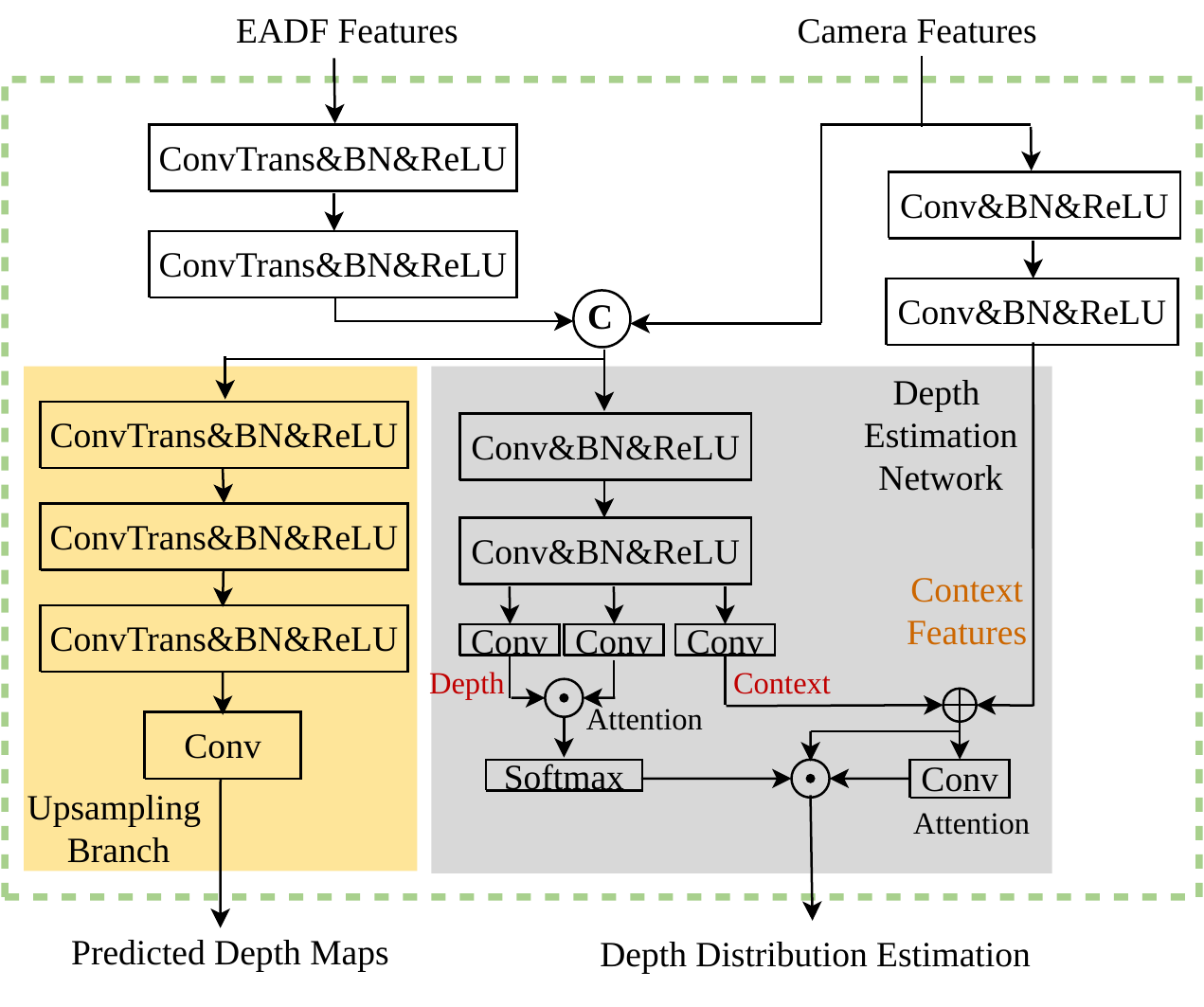}
\end{center}
   \caption{The detailed architecture of the EA-LSS framework.} 
\label{fig:depth_sem_arch}
\end{figure}

\begin{figure}[t]
\begin{center}
\includegraphics[width=\linewidth]{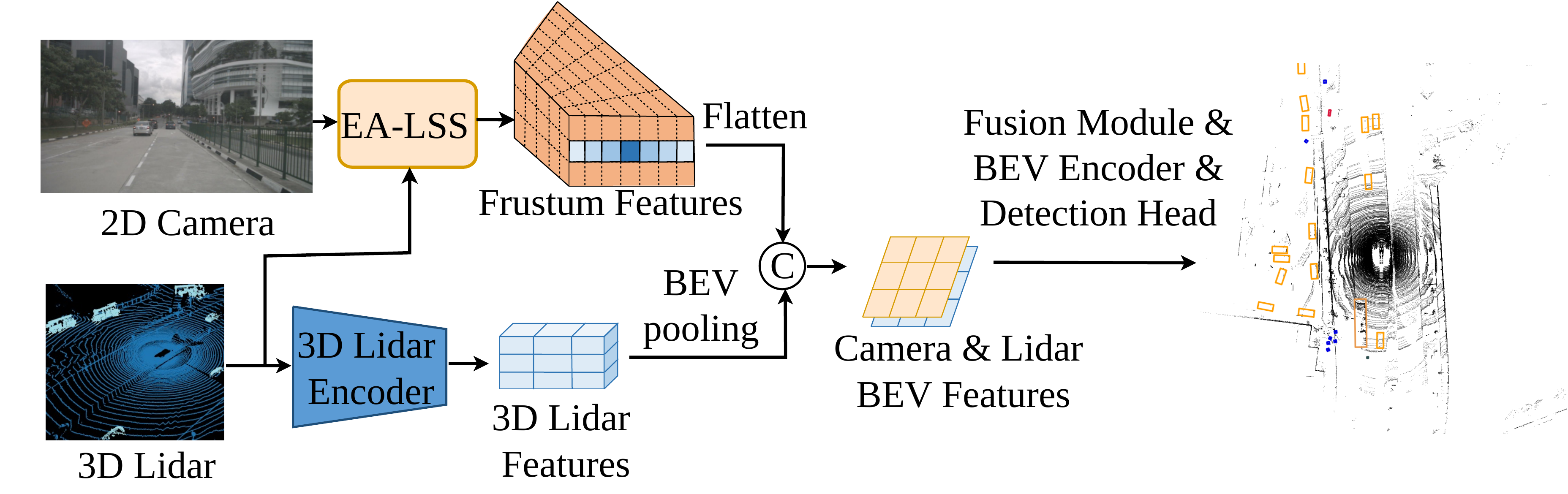}
\end{center}
   \caption{One sample of LSS-based 3D object detection models. The purpose of our proposed EA-LSS (shown in orange) is to transform camera feature map into 3D Ego-car coordinate. } 
\label{fig:bev_arch}
\end{figure}
\subsection{Edge-aware Lift-splat-shot Framework}
Our proposed EA-LSS framework is coupled by a fine-grained depth (FGD) module and an edge-aware depth fusion (EADF) module. The proposed framework is able to estimate the fine-grained global depth distribution while attending depth jump regions (i.e., the edges of objects). 

As shown in Figure \ref{fig:depth_sem_arch}, in order to fully utilize the depth information, output features of EADF module $\mathbf{F}^{EADF}$ are fed into convolutional layers to extract their geometric features. And geometric features are fused with image features as inputs to the depth network. 
Besides, we believe that some semantic information of the image might be lost after fusing the geometric information. Hence two convolutional layers with a skip connection are used to help the network to restore the lost semantic information.

As shown in Figure \ref{fig:bev_arch}, the proposed EA-LSS framework is plug-and-play and can be used in various LSS-based BEV 3D object detection methods~\cite{huang2021bevdet,liang2022bevfusion, CaDDN, li2022bevdepth}, for transforming camera feature map into the 3D Ego-car coordinate. 

The total loss $L_{total}$ is defined as: 
\begin{equation}
\label{eq:L_total}
L_{total} = L_{FGD}+L_{EADF}+L_{cls}+L_{box},
\end{equation}
where $L_{FGD}$ and $L_{EADF}$ represent FGD loss and EADF loss defined in Eqn~(\ref{eq:SDL}) and Eqn~(\ref{eq:GEDL}), respectively.  $L_{cls}$ and $L_{box}$ are classification loss and bounding box loss in the detection head.

\begin{table*}
\begin{center}
\small
\begin{tabular}{c|c|ccc}
\hline
Method & Modality & mAP  & NDS & Latency(ms)\\
\hline
BEVFormer~\cite{li2022bevformer} & C &  41.6  & 51.7 & -  \\
TransFusion~\cite{bai2022transfusion} & C+L  &	67.5 &	71.3 & 156.6 \\
DeepInteraction~\cite{yang2022deepinteraction} & C+L  &	69.9 &	72.6 &	204.1 \\
CMT~\cite{yan2023cmt} & C+L  &	70.3 &	72.9 &	- \\
\hline
BEVDepth-R50~\cite{li2022bevdepth} & C &	33.0 & 43.6 & 110.3 \\
+EA-LSS  & C  & 33.4 & 44.1 & 110.3 \\

Tig-bev~\cite{huang2022tigbev} & C &	33.8 & 37.5 & 68.0 \\
+EA-LSS  & C  & 35.9 & 40.7 & 68.0 \\
\hline
BEVFusion~\cite{liu2022bevfusion} & C+L  &	 68.5 & 71.4 & 119.2 \\
+EA-LSS & C+L  & 69.4 & 71.8 & 123.6 \\	

BEVFusion~\cite{liang2022bevfusion} & C+L &	69.6  & 72.1 & 190.3 \\
+EA-LSS  & C+L & \textbf{71.2}  &  \textbf{73.1} & 194.9 \\	
\hline
\end{tabular}
\end{center}
\caption{Comparison results of 3D object detection on nuScenes validation dataset. For fair comparison, 3D point cloud is not used in baselines BEVDepth and Tig-bev in the inference stage.}
\label{table:detection_val}
\end{table*}


\begin{table*}
\begin{center}
\small
\begin{tabular}{c|c|cc|ccccc}
\hline
Method & Modality & mAP$\uparrow$  & NDS$\uparrow$ & mATE$\downarrow$ & mASE$\downarrow$ & mAOE$\downarrow$ & mAVE$\downarrow$ & mAAE$\downarrow$ \\
\hline
BEVDet~\cite{huang2021bevdet} & C &  42.2  & 48.2 & 0.529&0.236&0.396&0.979&0.152  \\
BEVFormer~\cite{li2022bevformer} & C &44.5&53.5& 0.582& 0.256& 0.375& 0.378 & 0.126 \\
CenterPoint~\cite{yin2021centerpoint} & L &60.3& 67.3& 0.262 & 0.239 & 0.361 & 0.288& 0.136 \\
TransFusion~\cite{bai2022transfusion} & C+L &68.9& 71.6& 0.259 & 0.243 & 0.359 & 0.288& 0.127 \\
CMT~\cite{yan2023cmt}  & C+L &70.4& 73.0& 0.299 & 0.241 & 0.323 & 0.240& \textbf{0.112} \\
DeepInteraction~\cite{yang2022deepinteraction}  & C+L & 70.8 & 73.4 & 0.257 & 0.240 & 0.325 & 0.245 & 0.128 \\
\hline
BEVFusion~\cite{liang2022bevfusion} & C+L &	71.3  & 73.3 & 0.250& 0.240& 0.359& 0.254& 0.132 \\
+EA-LSS  & C+L & 72.2& 74.4& 0.247&0.237& 0.304& 0.250 & 0.133 \\	
+EA-LSS*  & C+L & \textbf{76.5} &  \textbf{77.6} & \textbf{0.233} & \textbf{0.228} & \textbf{0.281} & \textbf{0.196} & 0.123\\	
\hline
\end{tabular}
\end{center}
\caption{The comparison result of 3D object detection on nuScenes testing dataset. * represent the test time augment and model ensemble. Our EA-LSS reaches top1 in the leaderboard of nuScenes detection task.}
\label{table:detection_test}
\end{table*}

\section{Experiments}
\subsection{Implementation Details}
In our experiments, we use BEVFusion~\cite{liu2022bevfusion}/ BEVFusion~\cite{liang2022bevfusion}/ BEVDepth~\cite{li2022bevdepth}/Tig-bev~\cite{huang2022tigbev} as the baseline methods to verify the effectiveness and efficiency of our proposed EA-LSS framework. We follow the same training strategies as these baselines.

For fair comparison with camera-only LSS-based methods~\cite{li2022bevdepth,huang2022tigbev}, we use the same input as BEVDepth~\cite{li2022bevdepth} and Tig-bev~\cite{huang2022tigbev}, which means the 3D point cloud information is not directly used as the input. Only EADF Loss and FGD Loss are used to constrain depth estimation network during training stage.


The channel numbers of the two convolutional layers after EADF features are 32 and 64. The channel numbers of the three convolutional layers in upsampling branch are set to 256, 128 and 128. The kernel size of these convolutional layers is 5, the padding size is 2, the stride size is 2.

The channel numbers of the five convolutional layers in depth estimation network are set to 256, 256, 40, 128, 1. The channel numbers of the two convolutional layers before context features are set to 256 and 128. The kernel size of these convolutional layer is 3, the padding size is 1, the stride size is 1.

In the expanding operation of edge-aware depth fusion module, step size k is set to 7. The number of views $N_v$ is set to 6. Height and weight of depth maps $N_H$ and $N_W$ are set to 256 and 704.
Both $L_{FGD}$ and $L_{EADF}$ adopt focal Loss, and $\gamma$ and ${\alpha}_{c}$ are set to 2.0 and 0.25, respectively. Our training environment is on a server with 8*NVIDIA A100 GPU and AMD EPYC 7402 CPU.


\subsection{Dataset}
In order to evaluate the effectiveness of our proposed EA-LSS framework, we conduct comparison experiments on nuScenes Dataset~\cite{caesar2020nuScenes}. This dataset contains 40k labeled samples, with 23 different classes. We use mean Average Precision (mAP) and nuScenes detection score (NDS) as evaluation indicators. 
mAP and NDS on validation set are reported. Meanwhile, we also report the latency time to illustrate the negligible increase of inference time after using our proposed EA-LSS framework. 

\subsection{Comparison Results}

Table~\ref{table:detection_val} reports experiment results of nuScenes 3D object detection validation dataset. Our proposed EA-LSS framework can be added to any LSS-based baseline methods~\cite{li2022bevdepth,liu2022bevfusion, liang2022bevfusion,huang2022tigbev} with negligible increment of inference time. When using camera-only baseline method Tig-bev~\cite{huang2022tigbev}, mAP and NDS increase 2.1\% and 3.2\% after applying the proposed EA-LSS framework. 
On baseline method BEVFusion~\cite{liang2022bevfusion}, mAP and NDS increase 1.6\% and 1.0\%, respectively.

In Table~\ref{table:detection_test}, we report the experiment results of nuScenes 3D object detection test dataset. After using EA-LSS framework, the mAP and NDS for baseline BEVFusion~\cite{liang2022bevfusion} increase 0.9 \% and 1.1\%, respectively. After adding test time augment and model emsemble, the mAP and NDS of our proposed EA-LLS framework reach 76.5\% and 77.6\%, achieving the Top 1 in leaderboard of nuScenes Detection Task.

\begin{figure*}[t]
\begin{center}
\includegraphics[width=0.92\linewidth]{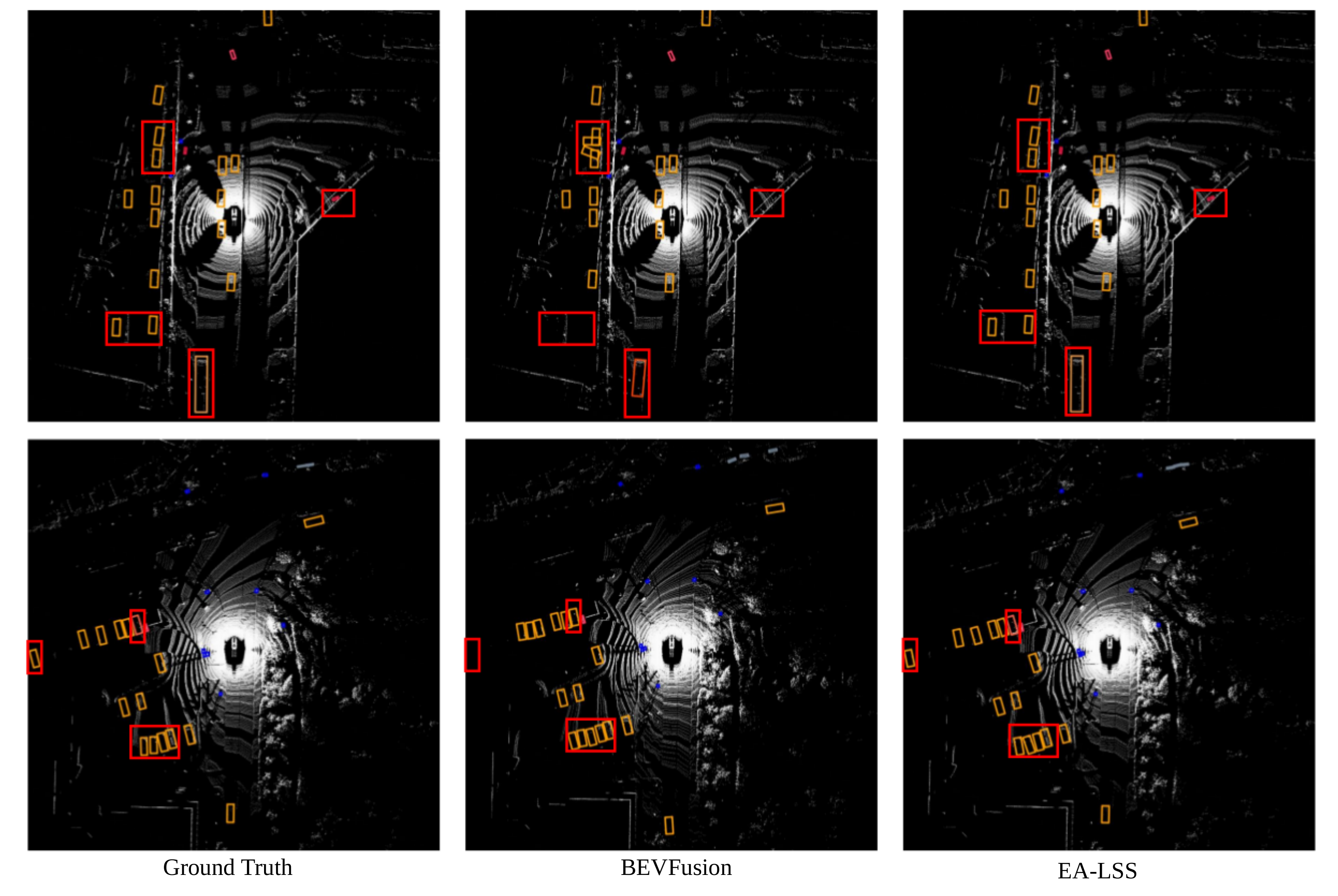}
\end{center}
   \caption{Visualization of the nuScenes 3D object detection validation dataset. From the left to the right, we provide results of ground-truth, baseline method BEVFusion~\cite{liu2022bevfusion} before and after adding our proposed EA-LSS. Results on cars are colored in yellow, pedestrian in blue and bicycle in pink. The red boxes highlight the difference before and after using our proposed EA-LSS. 
   We can observe that results after adding our proposed EA-LSS projector are better than before.  }
   \label{fig:detect_visual}
\end{figure*}

\begin{table}
\begin{center}
\begin{tabular}{cc|cc|cc}
\hline
\multicolumn{2}{c|}{FGD module} & \multicolumn{2}{c|}{EADF module} & mAP & NDS \\
 &  & C-DG' & EADFL  &  &  \\
\hline
\multicolumn{2}{c|}{ \usym{2715} } & \usym{2715} & \usym{2715} &  69.6  &	72.1  \\
\multicolumn{2}{c|}{ \usym{2713}} & \usym{2715} & \usym{2715} &	 70.1	 &	 72.4 \\
\multicolumn{2}{c|}{\usym{2713}} & \usym{2713}& \usym{2715} &	70.6	 &	72.8  \\
\multicolumn{2}{c|}{ \usym{2713}} & \usym{2713} & \usym{2713} & 71.2 &	73.1  \\
\multicolumn{4}{c|}{Improvment} & +1.6 & +1.0\\
\hline
\end{tabular}
\end{center}
\caption{Ablation study of our proposed FGD module and EADF module. Here, C-DG' represents concatenation of multi-view depth map $\mathbf{D}$ and multi-view edge maps $\mathbf{G'}$, while EADFL represents the usage of EADF loss.} 
\label{table:ablation}
\end{table}

\subsection{Ablation Study}

In Table~\ref{table:ablation}, we provide ablation study for different components in EA-LSS framework on nuScenes benchmarks. We use BEVFusion~\cite{liang2022bevfusion} as baseline method to evaluate the effectiveness of EADF module and FGD module.
After only adding FGD module, the mAP and NDS are improved for 0.5 \% and 0.3 \%, respectively. Next, after adding EADF module, the mAP and NDS continue to be improved for 1.1 \% and 0.7 \%, respectively.  We also conduct the ablation study on EADF Loss and concatenation of  $\mathbf{D}$ and $\mathbf{G'}$, and the mAP is improved for 0.5 \% and 0.6 \%, respectively.

\begin{table}
\begin{center}
\begin{tabular}{c|c|ccccccc}
\hline
$k$ &	mAP & NDS \\
\hline
3 &	68.9 &	71.6 \\
5 &	69.1 &	71.6  \\
7 &	\textbf{69.4} &	\textbf{71.8} \\
9 & 68.4 & 71.4 \\						
\hline
\end{tabular}
\end{center}
\caption{Performances on nuScenes 3D object detection validation dataset, when using different step $k$ of block division and expanding operation in edge-aware depth fusion module.}
\label{table:k}
\end{table}

Table~\ref{table:k} provides performances on nuScenes 3D object detection validation dataset, when using different step size $k$ of block division and expanding operation in EADF module. We use BEVFusion~\cite{liu2022bevfusion} as baseline method. The best mAP and NDS are achieved when $k=7$. 
A large step size leads to low resolution of the depth map, while a small step size leads to a large number of zero-valued points in the depth map.

Upperbound analysis is adopted to analyze the impact of wrong predictions in the ``depth jump" region by depth estimation networks. In Table~\ref{table:ceiling_analysis}, we use 3D Lidar information as the ground-truth (GT) to replace the predicted depth at scene edges for baseline BEVFusion~\cite{liu2022bevfusion}. The mAP and NDS of the camera branch using GT are improved by 3.6\% and 2.9\% compared to using the predicted depth, which demonstrates the importance of ``depth jump" problems.


\subsection{Qualitative Evaluation}
We provide visualization results in Figure~\ref{fig:detect_visual} to demonstrate the effectiveness of our proposed EA-LSS projector on nuScenes 3D object detection.
Experiment results of ground-truth, baseline method BEVFusion~\cite{liu2022bevfusion} and baseline method using our proposed EA-LSS framework are in the first column, second column and third column of Figure~\ref{fig:detect_visual}. The cars are shown in yellow, the pedestrian in blue and truck in red.  More target objects can be captured by adding our proposed EA-LSS projector than using the baseline method BEVFusion~\cite{liu2022bevfusion} only.

\begin{table}
\begin{center}
\begin{tabular}{c|c|c}
\hline
Method & mAP  & NDS \\
\hline
BEVFusion-C w/o GT~\cite{liu2022bevfusion} & 35.3 & 36.4 \\
BEVFusion-C w GT~\cite{liu2022bevfusion} &  38.9 & 39.3 \\
\hline
\end{tabular}
\end{center}
\caption{Upperbound analysis on nuScenes 3D object detection validation dataset. -C represents the camera branch.}
\label{table:ceiling_analysis}
\end{table}

\section{Conclusions}

In this paper, we propose a novel edge-aware lift-splat-shot (EA-LSS) framework, which is plug-and-play and can be used in any LSS-based methods. 
The proposed framework contains a edge-aware depth fusion (EADF) module and a fine-graind depth (FGD) module.
EADF module is proposed to alleviate the ``depth jump'' problem in depth estimation network by focusing on the edge of each object.
In order to capture more accurate depth distribution, our proposed FGD module can fully utilize the 3D Lidar depth information to supervise the depth estimate network, by using better size match between the predicted depth map and the GT depth map.
To validate effectiveness and efficiency, we conduct experiments on nuScenes 3D object detection benchmarks. Experiment results demonstrate that EA-LSS framework effectively improves the accuracy of camera-only and multi-modal BEV 3D object detection methods with negligible increase of inference time. 

\bibliography{aaai24}

\end{document}